\documentclass[sigconf]{acmart}
\AtBeginDocument{%
  }

\copyrightyear{2026}
\acmYear{2026}
\setcopyright{cc}
\setcctype{by}
\acmConference[KDD '26]{Proceedings of the 32nd ACM SIGKDD Conference on Knowledge Discovery and Data Mining V.2}{August 09--13, 2026}{Jeju Island, Republic of Korea}
\acmBooktitle{Proceedings of the 32nd ACM SIGKDD Conference on Knowledge Discovery and Data Mining V.2 (KDD '26), August 09--13, 2026, Jeju Island, Republic of Korea}
\acmDOI{10.1145/3770855.3818886}
\acmISBN{979-8-4007-2259-2/2026/08}

\usepackage{listings}         
\usepackage{cleveref}         
\usepackage{float}            
\usepackage{subcaption}       
\usepackage{booktabs}  
\usepackage{multirow}  
\usepackage{nicefrac}         

\definecolor{KleinBlue}{RGB}{0,47,167}
\definecolor{TiffanyBlue}{RGB}{129,216,207}      
\definecolor{VeryPeri}{RGB}{102,103,171}
\definecolor{DeepTaupe}{RGB}{124,94,96}
\definecolor{WhiteSand}{RGB}{236,233,228}
\definecolor{VolcanicGlass}{RGB}{97,92,97}

\definecolor{LikePurple}{RGB}{138,43,226}        
\definecolor{LikeOrange}{RGB}{245,195,134}       
\definecolor{LikeBlue}{RGB}{156,188,227}         
\definecolor{LikeGreen}{RGB}{177,202,162}        

\definecolor{8A2BE2}{RGB}{138,43,226}
\definecolor{F5C386}{RGB}{245,195,134}
\definecolor{9CBCE3}{RGB}{156,188,227}
\definecolor{B1CAA2}{RGB}{177,202,162}

\begin{document}

\title{Mask-Proof: An LLM-based Automated Data Curation Pipeline on Mathematical Proofs}


\author{Jierui Zhang}
\authornote{Both authors contributed equally to this research.}
\email{2024140911@bupt.cn}
\affiliation{%
  \department{School of Computer Science}
  \institution{Beijing University of Posts and Telecommunications}
  \city{Beijing}
  \country{China}
}

\author{Siyuan Tan}
\authornotemark[1]
\email{cream_milk@bupt.edu.cn}
\affiliation{%
  \department{Graduate College for Engineers}
  \institution{Beijing University of Posts and Telecommunications}
  \city{Beijing}
  \country{China}
}

\author{Xinhang Li}
\email{24110180028@m.fudan.edu.cn}
\affiliation{%
  \department{School of Mathematical Sciences}
  \institution{Fudan University}
  \city{Shanghai}
  \country{China}
}

\author{Longzhuangzhi Lin}
\email{linlong0591@bupt.edu.cn}
\affiliation{%
  \department{School of Cyberspace Security}
  \institution{Beijing University of Posts and Telecommunications}
  \city{Beijing}
  \country{China}
}

\author{Dailin Li}
\email{ldlbest@mail.dlut.edu.cn}
\affiliation{%
  \department{School of Computer Science and Technology}
  \institution{Dalian University of Technology}
  \city{Dalian}
  \country{China}
}

\author{Chengfeng Gu}
\email{cgfa9563@gmail.com}
\affiliation{%
  \department{Chu Kochen Honors College}
  \institution{Zhejiang University}
  \city{Hangzhou}
  \country{China}
}

\author{Xinping Li}
\email{rccp@tsinghua.edu.cn}
\affiliation{%
  \department{Department of Psychological and Cognitive Sciences}
  \institution{Tsinghua University}
  \city{Beijing}
  \country{China}
}

\author{Yaxian Hao}
\email{haoyaxian@bupt.edu.cn}
\affiliation{%
  \department{Graduate College for Engineers}
  \institution{Beijing University of Posts and Telecommunications}
  \city{Beijing}
  \country{China}
}

\author{Shengjia Liang}
\email{liangshengjia@buaa.edu.cn}
\affiliation{%
  \department{State Key Laboratory of Virtual Reality Technology and Systems}
  \institution{Beihang University}
  \city{Beijing}
  \country{China}
}

\author{Yuxiang Ren}
\authornote{Corresponding author.}
\email{renyuxiang@nju.edu.cn}
\affiliation{%
  \department{School of Intelligence Science and Technology}
  \institution{Nanjing University}
  \city{Nanjing}
  \country{China}
}

\author{Wenhao Liu}
\authornotemark[2]
\email{wenhaoliu@bupt.edu.cn}
\affiliation{%
  \department{School of Computer Science}
  \institution{Beijing University of Posts and Telecommunications}
  \city{Beijing}
  \country{China}
}
\renewcommand{\shortauthors}{Jierui Zhang et al.}

\begin{abstract}
Large language models (LLMs) are increasingly capable of mathematical problem solving and can even assist with research-level proofs, yet we still lack a scalable and reproducible way to measure step-level reasoning in long proofs across diverse sources. This evaluation gap limits trustworthy AI assistance in proof-certified scientific progress. Existing evaluations often emphasize final answers or rely on costly expert grading, while end-to-end proof generation remains open-ended and hard to verify automatically.
We introduce \textbf{Mask-Proof}, a pipeline that turns real proofs into automatically checkable masked-step tasks. It masks key formula steps, provides the necessary surrounding context, and evaluates model reconstructions with an LLM-based equivalence judge using repeated votes for stability.
The resulting \textbf{Mask-ProofBench} contains 292 curated problems across diverse research areas. Experiments with 17 models show that reasoning-enhanced models outperform standard models by 12\% to 27\%. Our evaluator achieves 96.8\% agreement with expert annotators, enabling faithful, reproducible, and comparable measurement of step-level mathematical reasoning. Benchmark, annotations, and code are available at \url{https://github.com/weating/Mask-Proof}.
\end{abstract}

\ccsdesc[500]{Computing methodologies~Artificial intelligence}
\ccsdesc[300]{Computing methodologies~Natural language processing}

\keywords{Mathematical Reasoning, Large Language Models, Benchmark, Proof Evaluation, Automated Curation}


\maketitle

\section{Introduction}
\label{sec:introduction}
Large language models (LLMs) are increasingly competent at mathematical problem solving~\cite{balunović2026matharenaevaluatingllmsuncontaminated} and can even assist with research-level proofs~\cite{NEURIPS2022_9d560961,trinh2024solving}. Yet we still lack a scalable and reproducible measurement instrument for step-level reasoning in long-range proof—one that remains faithful to the underlying mathematics across heterogeneous sources such as research papers and competition benchmarks~\cite{ma2025reliablefinegrainedevaluationnatural,zheng-etal-2025-processbench, pandit2025hard2verifysteplevelverificationbenchmark}. This gap is increasingly visible in recent frontier-system investigations. For instance, Sébastien Bubeck and collaborators at OpenAI report that GPT-5 can sometimes make meaningful progress on research-level proof tasks under expert guidance~\cite{bubeck2025earlyscienceaccelerationexperiments}. However, such evidence often relies on step-wise expert judgment, which is costly to scale, difficult to reproduce, and hard to compare across domains.

\paragraph{Why This Matters for AI for Sciences.} This measurement gap is not merely a benchmarking nuisance. Mathematics is a foundational scientific discipline in which progress is certified by proofs. Accordingly, trustworthy AI assistance in mathematics requires domain-grounded evaluation of reasoning processes, not only final outcomes~\cite{chen2022program, ma2025reliablefinegrainedevaluationnatural, yang2025utmathmathevaluationunit}. Mask-Proof contributes proof-centric data and evaluation tools that make step-level reasoning measurable, reproducible, and comparable across real-world proof sources.

\paragraph{The Measurement Problem in Step-Level Proof Evaluation.} Treating step-level evaluation as a scientific measurement reveals three intertwined obstacles: what to measure, what context is required, and how to score correctness.
(1) \emph{Step selection is non-trivial.} Proofs vary widely in style and granularity. Many lines are trivial, purely definitional, or dominated by local surface cues. Masking such steps risks measuring shortcut exploitation rather than genuine inference, undermining construct validity~\cite{Geirhos_2020, zhu2023dyval, yu2023metamath}.
(2) \emph{Dependencies are often missing.} Research proofs frequently rely on external dependencies—cross-referenced lemmas, macros, or unstated background facts. If these dependencies are not recovered, a masked step may be underdetermined from the provided context, confounding reasoning ability with missing information and making failures uninterpretable.
(3) \emph{Scalable checking is difficult.} Even when a ground-truth step is available, correctness checking at scale remains hard: string matching is inadequate, symbolic tools offer no general guarantees, and expert evaluation does not scale. Automated judging therefore requires explicit mechanisms to control variance and ensure reproducibility~\cite{kim2023prometheus, thakur2025judgingjudgesevaluatingalignment, ke2024critiquellminformativecritiquegeneration, zhengJudgingLLMasJudge}. Taken together, these challenges prevent step-level reasoning in mathematics from being studied as a reproducible scientific phenomenon.

\begin{figure*}[t]
    \centering
    \includegraphics[width=\textwidth]{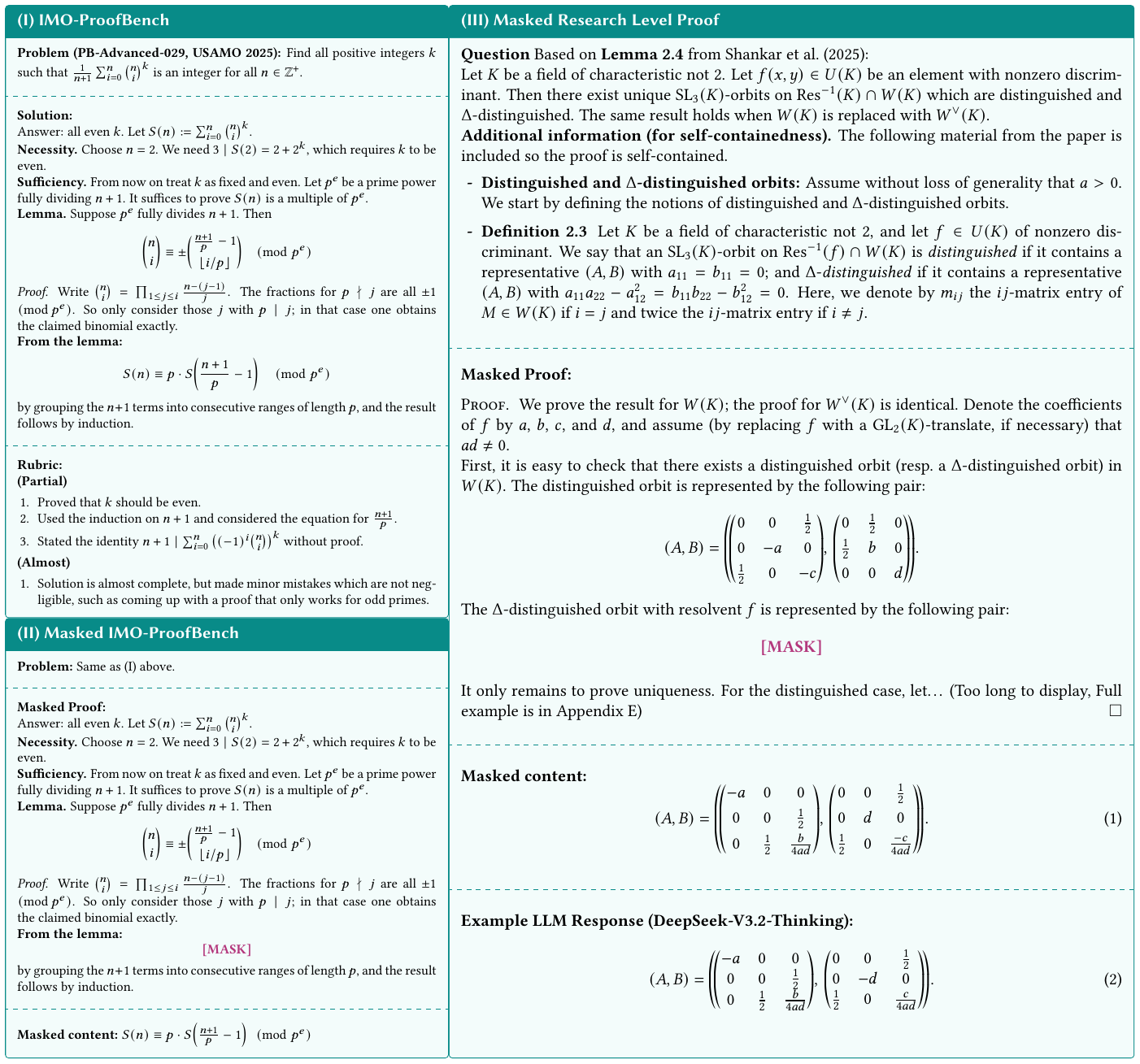} 
    \caption{
        (I) IMO-ProofBench~\cite{luong-etal-2025-towards} represents competition-level proof problems paired with full solutions and rubric-style evaluation.
        (II) After Mask-Proof processing, the same proof data are converted into a verifiable masked format, where key formula/derivation steps are hidden as targets for automatic evaluation.
        (III) Mask-ProofBench extends this masked, verifiable format to research-level proofs with substantially longer derivations, masking rigorous intermediate steps rather than final numerical results or highly templated fragments.
        See Appendix~\ref{app:artanecase} for complete details.
    }
    \Description[Three-panel comparison of proof evaluation formats]{Panel I shows an IMO-ProofBench competition problem (USAMO 2025) with a full solution involving binomial coefficient identities and a partial-credit rubric. Panel II shows the same proof after Mask-Proof processing, where a key congruence formula is replaced with a MASK token while the surrounding proof context is preserved. Panel III shows a research-level masked proof based on Lemma 2.4 from Shankar et al., involving SL3 orbit classification over fields of characteristic not 2, with additional self-containedness information including Definition 2.3, a masked matrix construction step, and an example LLM response from DeepSeek-V3.2-Thinking that incorrectly inverts the sign of the (2,2) entry.}
    \label{fig:mainfig}
\end{figure*}

\paragraph{Mask-Proof: Making Step-Level Reasoning Measurable.}
We introduce \textbf{Mask-Proof}, a pipeline that directly addresses the obstacles above by transforming heterogeneous proofs into verifiable, self-contained, step-level evaluation problems with variance-controlled automatic checking. The core idea is intentionally simple: rather than scoring entire proofs end-to-end, we test whether a model can reconstruct a key formula step whose ground truth is extracted directly from the original proof. Concretely, Mask-Proof (i) uses Codex CLI to agentically select key steps that cannot be solved by local cues alone, (ii) recovers the minimal dependency closure needed to make each masked step self-contained, meaning that no external lemmas, definitions, or macros are further required beyond the provided context, and (iii) checks model outputs using a calibrated LLM-based equivalence judge, with repeated independent judgments and majority voting to reduce variance. Beyond benchmarking, Mask-Proof serves as a reusable measurement framework, exhibiting effectiveness, reproducibility, and generalization across proof sources. Figure~\ref{fig:mainfig} illustrates examples of the resulting masked-step format. 

In this work, we make three contributions:
\begin{enumerate}
    \item \textbf{Mask-Proof}, an LLM-based automated curation pipeline that identifies inference-critical formula steps, reconstructs self-contained proof context, and produces verifiable masked-step problems from real-world mathematical proofs.
    \item \textbf{Mask-ProofJudge}, an automatic step-level evaluation framework calibrated against expert judgments, achieving 96.8\% human--judge agreement with variance reduction via repeated independent judgments and voting.
    \item \textbf{Mask-ProofBench}, a benchmark generated by this pipeline that enables scalable and reliable evaluation of step-level reasoning in real-world research-level mathematical proofs. This benchmark contains 292 Mask-Proof problems, all of which were manually audited by human experts in mathematics.
\end{enumerate}

\section{Related Work}
\label{sec:related_work}
\paragraph{Answer-based mathematical reasoning benchmarks.}
Evaluation of LLM mathematical reasoning has historically centered on benchmarks with uniquely verifiable endpoints. 
GSM8K~\cite{cobbe2021gsm8k} and MATH~\cite{hendrycksmath2021} established final-answer accuracy as the dominant metric, enabling scalable evaluation and providing limited visibility into reasoning processes.
As performance on these benchmarks saturated, subsequent work increased difficulty through olympiad-level problems~\cite{he2024olympiadbench,gao2024omnimathuniversalolympiadlevel}, expert-authored frontier evaluations~\cite{glazer2024frontiermath}, and research-level question-answering~\cite{zhang2025realmath}.
However, even sophisticated answer-based benchmarks fundamentally evaluate \emph{conclusions} rather than \emph{formulas}, leaving proof quality unexamined.

\paragraph{Proof-centric evaluation and formal theorem proving.}
Proof-centric evaluation shifts focus from answer correctness to formula rigor, making verifiability the core challenge.
Formal theorem proving addresses this by delegating correctness to proof assistant kernels:
miniF2F~\cite{zheng2021minif2f} and PutnamBench~\cite{tsoukalas2024putnambench} benchmark formal proving on competition mathematics.
However, the natural-language-to-formal gap introduces semantic errors that are difficult to eliminate;
ReForm~\cite{chen2025reform} reveals that autoformalization is inherently challenging, with even human experts producing semantic errors in 38.5\% of cases.
Natural-language proof evaluation is closer to real mathematical practice but substantially harder to grade at scale.
IMO-ProofBench~\cite{luong-etal-2025-towards} introduces rubric-based grading for competition proofs, while STORM-BORN~\cite{liu-etal-2025-storm} curates challenging formulas from research papers with human-in-the-loop verification.
More recently, IMProofBench~\cite{schmitt2025improofbenchbenchmarkingairesearchlevel} targets \emph{research-level} proof generation with expert-authored problems, and combines expert grading of full proofs with  gradable follow-up subquestions. Yet these evaluations still require substantial human expert annotation and can't scale to evaluate in-the-wild research-level proofs.

\paragraph{Verifiers and scalable checking.}
To reduce reliance on expert grading, an important line of work trains verifiers that score reasoning processes.
Process reward models~\cite{lightman2023verify,luo2024automatedprocesssupervision} provide step-level feedback, while diagnostic studies reveal that LLMs struggle to locate reasoning errors but can correct them given error locations~\cite{tyen-etal-2024-llms}.
Self-correction approaches have shown limited success: Huang et al.~\cite{huang2024selfcorrect} demonstrate that LLMs cannot reliably self-correct reasoning without external feedback.
More recent generator--verifier frameworks, such as DeepSeekMath-V2~\cite{shao2025deepseekmathv2selfverifiablemathematicalreasoning}, scale verification through meta-verification and self-training, though their evaluation remains centered on competition-style problems.
High-profile demonstrations of AI-assisted mathematics~\cite{bubeck2025earlyscienceaccelerationexperiments} still emphasize expert oversight rather than fully automatic evaluation.
Collectively, prior work highlights a persistent capability--verifiability tension: as evaluations approach research-level proofs, obtaining low-cost, scalable, and reproducible verification becomes substantially harder.

\begin{figure*}[t]
    \centering
    \includegraphics[width=\textwidth]{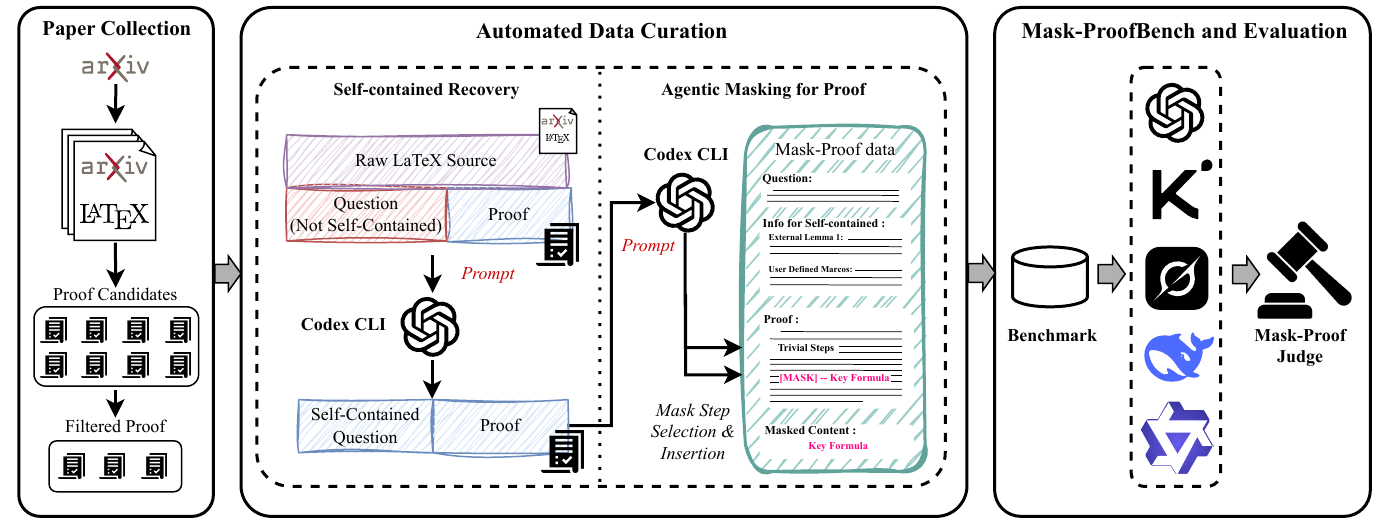}
    \caption{\textbf{Mask-Proof Pipeline.}
Starting from raw arXiv LaTeX sources, our pipeline extracts complete proofs, repairs them into self-contained contexts, and agentically masks a key formula step to produce Mask-ProofBench.}
    \Description[Mask-Proof pipeline overview diagram]{The pipeline has three stages. Stage one, Paper Collection: arXiv papers with LaTeX sources are downloaded and filtered into proof candidates. Stage two, Automated Data Curation, contains two sub-modules: Self-contained Recovery, where Codex CLI takes a non-self-contained question and raw LaTeX source and resolves dependencies to produce a self-contained question paired with the proof; and Agentic Masking for Proof, where Codex CLI selects a key formula step, replaces it with a MASK token, and records the masked content as ground truth, while trivial steps are filtered out. Stage three, Mask-ProofBench and Evaluation: the resulting benchmark instances are evaluated by multiple LLMs, and a Mask-Proof Judge assesses semantic equivalence between model outputs and the ground-truth masked content.}
    \label{fig:pipelineoverview}
\end{figure*}


\section{Methodology}

In this section, we describe the overall pipeline, Mask-Proof, including automated paper collection and proof quality control, automated proof data curation and evaluation. This LLM-based pipeline is explicitly designed to be contamination-resistant by collecting and curating recent arXiv submissions, reproducible by using powerful Codex CLI, and broadly generalizable by adding different sources of proof data, while ensuring that the resulting masked proof data supports fully automated verification and maintains rigorous guarantees of correctness and verifiability.
An overview of the pipeline is shown in Figure~\ref{fig:pipelineoverview}.

\subsection{Paper Collection and Proof Extraction}

To ensure the quality of the collected arXiv papers, proofs, and the resulting masked proof data, we curate a corpus of real-world, research-level mathematical papers under the following criteria. 
First, we restrict our corpus to arXiv submissions between August and October 2025, so as to minimize potential contamination from pretraining data and reduce the risk of data leakage in large language models. 
Second, we collect papers across a broad range of mathematical fields to mitigate domain bias; the distribution of the major subject categories is summarized in Table~\ref{tab:datadistributionpreview}.

We mine these research-level papers together with their full LaTeX sources to enable precise proof extraction and dependency resolution. 
To ensure that the resulting evaluation instances require nontrivial reasoning, we further filter out trivial proofs, including very short arguments, purely definitional proofs, and pattern-completable formulas, as such instances either do not involve substantive inference or are prone to context leakage.


\paragraph{Human verification of benchmark quality.}
Before downstream evaluation, all 292 final \textbf{Mask-ProofBench} instances were manually audited by domain experts in mathematics. 
The audit assessed the mathematical correctness of the masked ground-truth step within the original proof, the sufficiency of the recovered context for solving the masked step without external documents, the completeness of the selected mask as a formula-level reasoning step rather than a syntactic fragment, and screened each instance for obvious shortcut cues, malformed LaTeX, or unresolved paper-specific notation.
Instances that failed these checks were discarded; the expert verification annotations will be released with the benchmark.

\subsection{Automated Proof Data Curation}
\label{sec:mask_pipeline}

\paragraph{Self-contained Recovery}
Formally, we define a proof to be self-contained if the pipeline can recover a sufficient dependency closure such that each masked step is solvable from the extracted theorem statement and the reconstructed proof context alone, with no reliance on external documents.

Concretely, the pipeline provides Codex CLI with (a) the proof snippet and (b) its source LaTeX data, and instructs it to resolve \textbf{explicit dependencies} by traversing internal cross-references with explicit labels (e.g., \texttt{\textbackslash ref}, \texttt{\textbackslash eqref}), locating the corresponding labeled targets in the source, appending the extracted content to the problem context, and expanding user-defined macros that carry essential semantics into standard LaTeX. In contrast, \textbf{implicit dependencies} are not directly anchored by labels (e.g., symbols whose closed-form definitions appear elsewhere); these are recovered by searching the source LaTeX for their defining statements and attaching the extracted content as Additional Information. All relevant prompts are released with the code.
Cases dominated by external dependencies (e.g., citations to external conclusions) are discarded.

\paragraph{Agentic Masking for Proof}
For each self-contained proof, we construct masked steps by replacing exactly one contiguous reasoning step with \texttt{[MASK]}. Each masked step contains only pure mathematical expressions (no natural-language text). From each source proof, Mask-Proof generates up to three such masked steps, each corresponding to an independent and well-defined inference.
We defined a series of rubrics (see Appendix~\ref{app:maskrubric}) to guide the selection of masked positions. These rubrics instruct Codex CLI to agentically mask key inference-critical steps, ensuring that the resulting tasks necessarily require non-trivial mathematical reasoning rather than shallow pattern matching.

\paragraph{Faithful Mask Application}
A practical failure mode of directly applying LLM-based masking is unintended proof rewriting, which can subtly alter semantics. To preserve faithfulness to the original proof, our pipeline decouples \emph{(i) selecting mask steps} from \emph{(ii) applying the mask}. We use Codex CLI to perform these two steps automatically under strict constraints, ensuring that only the target step is masked while the remaining proof text remains unchanged.


\subsection{Automated Proof Data Evaluation}
\label{sec:auto_judger}

As illustrated in Figure~\ref{fig:pipelineoverview}, our automated proof-data curation pipeline converts each evaluation unit into a masked key formula step within a well-defined, self-contained proof context.

Compared to evaluating full proofs or sub-proofs, focusing on such mask instances substantially reduces ambiguity and leads to more consistent automated judgments.

Nevertheless, exact string matching remains inadequate for mathematical proofs due to extensive notational diversity and equivalent reformulations. To address this limitation, we introduce Mask-ProofJudge, an LLM-based evaluation framework where GPT-OSS-120B (High) serves as the backbone to assess semantic equivalence between model outputs and the ground-truth masked content.

To further reduce evaluation variance arising from generation stochasticity, we sample four solutions per problem and aggregate multiple independent judgments via majority voting, reporting the resulting average accuracy.

In practice, the LLM judge achieves a 96.8\% agreement rate with human expert annotations, indicating that the proposed evaluation protocol provides a reliable and high-confidence approach for expert-level assessment.

\subsection{Case Study}
\label{sec:case_study}

To explicitly demonstrate the workflow described in Sections~\ref{sec:mask_pipeline} and~\ref{sec:auto_judger}, we trace the processing of \textit{Lemma~2.4} from Shankar et al.~\cite{shankar2025counting}. This example, which corresponds to the case illustrated in Figure~\ref{fig:mainfig}, shows how each stage of our pipeline transforms a raw arXiv fragment into a rigorous evaluation instance.

\textbf{Stage 1: Self-contained Recovery.} 
In this case, the raw extraction of \textit{Lemma 2.4} requires constructing a matrix representation for the "$\Delta$-distinguished orbit." However, the critical definition of "$\Delta$-distinguished" is located in \textit{Definition 2.3}, outside the local context~\cite{shankar2025counting}. As defined in our curation protocol, this constitutes an "unresolved external dependency."
Our pipeline's dependency agent successfully identified this gap. Instead of discarding the proof, it traced the term back to \textit{Definition 2.3}, and appended it to the problem context. This step ensured the instance satisfied the self-contained criterion.

\textbf{Stage 2: Agentic Mask Selection.} 
Following the recovery, the Mask Selection Agent scanned the proof to identify a non-trivial reasoning step based on the rubrics in Appendix~\ref{app:maskrubric}.
The agent rejected steps involving simple algebraic rearrangements (violating the "non-trivial" rubric) and selected the specific matrix construction step.
This selection was strategic: correctly masking this matrix forces the model to synthesize the recovered global definition with local variables $a, b, c, d$. It ensures the task tests reasoning necessity rather than pattern matching or context copying.

\textbf{Stage 3: Automated Evaluation.} 
Finally, we evaluated the response from DeepSeek-V3.2-Thinking using Mask-ProofJudge.
The model's output visually resembled the ground truth but inverted the sign of the $(2,2)$ entry from $d$ to $-d$. Under the defining convention \(\mathrm{Res}(A,B)=4\det(xA-yB)\), the ground-truth representative yields \(f(x,y)\). The model response changes the signs of the \(B_{22}\) and \(B_{33}\) entries, yielding \(f(x,-y)\) instead of \(f(x,y)\). Thus the response does not lie in the required fiber \(\mathrm{Res}^{-1}(f)\), so the error is mathematical.

\section{Experiments}
\label{sec:experiments}

In this section, we evaluate Mask-Proof to ascertain its effectiveness in automated constructing masked problem following four questions:
RQ1: How effective is Mask-Proof as an automated data curation pipeline?
RQ2: How reproducible is the automated Mask-Proof pipeline?  
RQ3: Is our automated evaluation framework consistent with human experts in masked-step evaluation? 
RQ4: Can the Mask-Proof pipeline generalize to other data sources?

\begin{figure*}[t]
    \centering
    \includegraphics[width=\textwidth]{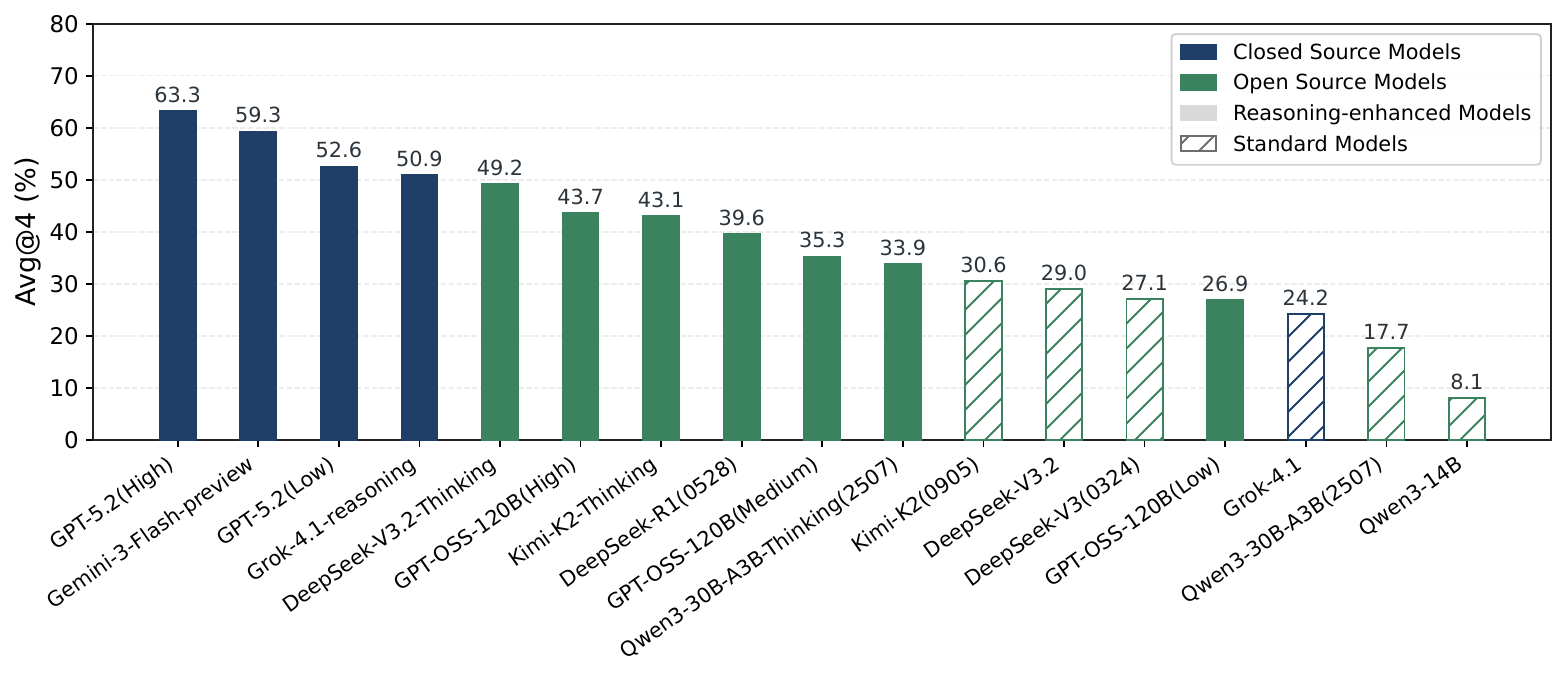}
    \caption{\textbf{Avg@4 leaderboard on Mask-ProofBench.}
    All 17 evaluated models are ranked by Avg@4 accuracy. Reasoning-enhanced models demonstrate a substantial performance advantage over standard models at comparable parameter scales.}
    \Description[Bar chart showing Avg@4 leaderboard on Mask-ProofBench]{A horizontal bar chart ranking 17 models by Avg@4 accuracy on Mask-ProofBench. GPT-5.2 High leads at 63.3 percent, followed by Gemini-3-Flash-preview at 59.3 percent, GPT-5.2 Low at 52.6 percent, Grok-4.1-reasoning at 50.9 percent, and DeepSeek-V3.2-Thinking at 49.2 percent. Standard models score below 30 percent, with Qwen3-14B lowest at 8.1 percent. Bars are color-coded by closed-source versus open-source and reasoning-enhanced versus standard, showing a clear performance boundary near 32 percent separating the two reasoning categories.}
\end{figure*}

\subsection{Experimental Setup}
\label{sec:exp_setup}

\paragraph{Dataset.}
To comprehensively evaluate the effectiveness of the Mask-Proof pipeline, we curate two datasets:
\begin{itemize}
    \item \textbf{Mask-ProofBench}, constructed using the pipeline described in Section~\ref{sec:mask_pipeline}. We collected 403 arXiv papers from August to October 2025, extracted 835 theorem--proof blocks, and retained 292 final masked instances.
    \item To evaluate the cross-dataset generalization of the automated pipeline, we additionally construct a comparison dataset by applying Mask-Proof to all 30 \textit{PB-Advanced} proofs from IMO-ProofBench~\cite{luong-etal-2025-towards}. This comparison dataset consists of 35 mask instances.
\end{itemize}

\paragraph{Evaluation Metrics.}

We evaluate the Mask-Proof pipeline and the curated dataset using three complementary metrics: Avg@4, Spearman’s rank correlation coefficient, and BLEU.
Avg@4 measures the average correctness of an LLM over four independent attempts on each masked proof step.
Spearman’s rank correlation coefficient assesses the transferability of Mask-Proof by quantifying the consistency of model rankings of IMO-ProofBench under masked evaluation.
BLEU evaluates n-gram overlap to measure surface-level linguistic consistency, and is used to assess the reproducibility of self-contained proof recovery.
Together, these metrics characterize the effectiveness, transferability, and reproducibility of the Mask-Proof pipeline.

\paragraph{Evaluated Models.}
To conduct a comprehensive evaluation, we categorize state-of-the-art LLMs into two groups based on their reasoning capability.

\begin{itemize}
    \item \textbf{Standard models:} General-purpose models that generate answers directly without explicit deliberation (e.g., DeepSeek-V3.2, Grok-4.1, Qwen3-14B), serving as baselines to quantify the benefit of test-time reasoning.
    \item \textbf{Reasoning-enhanced models:} Models enhanced for multi-step thinking (e.g., DeepSeek-R1, Grok-4.1-reasoning, Gemini-3-Flash-preview), which produce internal chains of thought before outputting the final answer.
\end{itemize}



\subsection{Effectiveness of the Automated Mask-Proof Pipeline (RQ1)}

\subsubsection{Main Results on Mask-ProofBench}
We evaluate 17 models on \textbf{Mask-ProofBench}, including both reasoning and non-reasoning categories. Figure~\ref{fig:avg_accuracy} presents the full leaderboard.

\paragraph{Reasoning Capability Creates a Qualitative Boundary.}
The results indicate that Mask-ProofBench clearly separates reasoning-enhanced models from standard models.
A performance boundary emerges at approximately 32\%: eight out of nine reasoning-enhanced models (except GPT-OSS-120B under low reasoning effort) exceed this threshold, whereas all six standard models fall below it.
To further quantify this effect, we conduct paired comparisons within the same model families. For example, Grok-4.1-reasoning outperforms Grok-4.1 by 26.7\%, demonstrating that the benchmark reliably and effectively captures improvements attributable to enhanced reasoning capability.

\paragraph{Highly discriminative, not saturated, and extensible.} 
The leaderboard exhibits a wide performance range, from a lower bound of 8.1\% (Qwen3-14B) to an upper bound of 63.3\% (GPT-5.2 High) in Avg@4, indicating strong discriminative power.
Notably, these results are obtained under a single-mask setting with our pipeline, where only one formula-level step is masked per proof.
This design leaves substantial headroom for increasing benchmark difficulty, for example by masking multiple interdependent steps or requiring multi-step reconstruction as model capabilities advance.

\paragraph{No Anomalous Reversals.}
As a sanity check, we verify that the ranking contains no major surprises: smaller models do not outperform larger ones, and models known to struggle with mathematics do not unexpectedly excel. This coherence indicates that the benchmark measures a stable capability rather than task-specific artifacts.

\begin{figure}[t]
    \centering
    \includegraphics[width=\columnwidth]{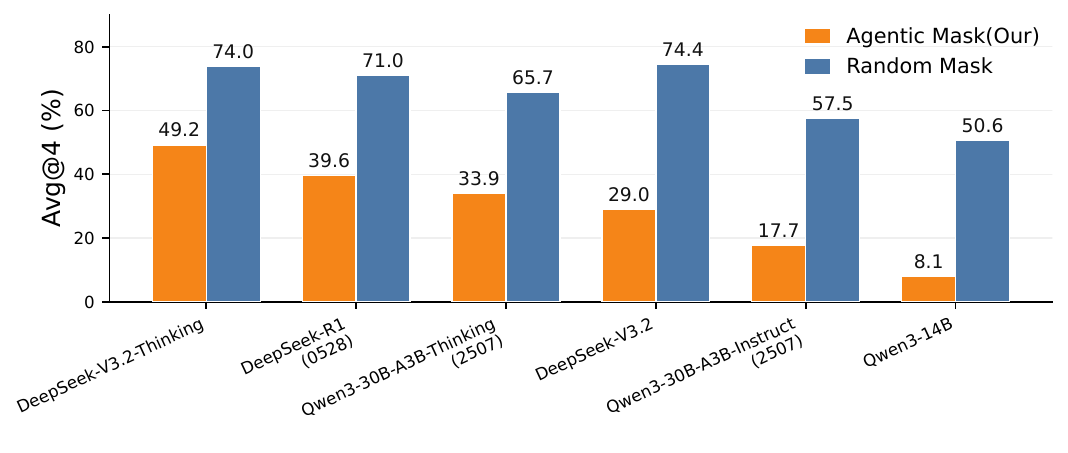}
    \caption{\textbf{Leaderboard distortion from random masking.} 
    Avg@4 performance under \emph{random} and \emph{strategic} masking for representative models. 
    Random masking substantially inflates scores—especially for standard models—while agentic masking consistently lowers performance and restores meaningful separation, demonstrating the necessity of anti-hack filtering for a discriminative benchmark.}
    \Description[Grouped bar chart comparing agentic and random masking]{A grouped bar chart comparing Avg@4 performance of six models under agentic masking versus random masking. Under agentic masking, scores range from 8.1 percent for Qwen3-14B to 49.2 percent for DeepSeek-V3.2-Thinking. Under random masking, all scores inflate substantially: DeepSeek-V3.2 rises from 29.0 to 74.4 percent, and Qwen3-14B from 8.1 to 50.6 percent. The performance gap between reasoning-enhanced and standard models is clearly visible under agentic masking but largely disappears under random masking.}
    \label{fig:distortion_scatter}
\end{figure}

\subsubsection{Ablation Study: Effectiveness of the Self-Contained Module}

\label{sec:selfcontained}
The Mask-Proof pipeline intentionally resolves cross-references and expands user-defined macros.
Manual inspection of all 292 masked instances in Mask-ProofBench confirms that 291 out of 292 are fully self-contained after this preprocessing.

\begin{figure}[t]
    \centering
    \includegraphics[width=\columnwidth]{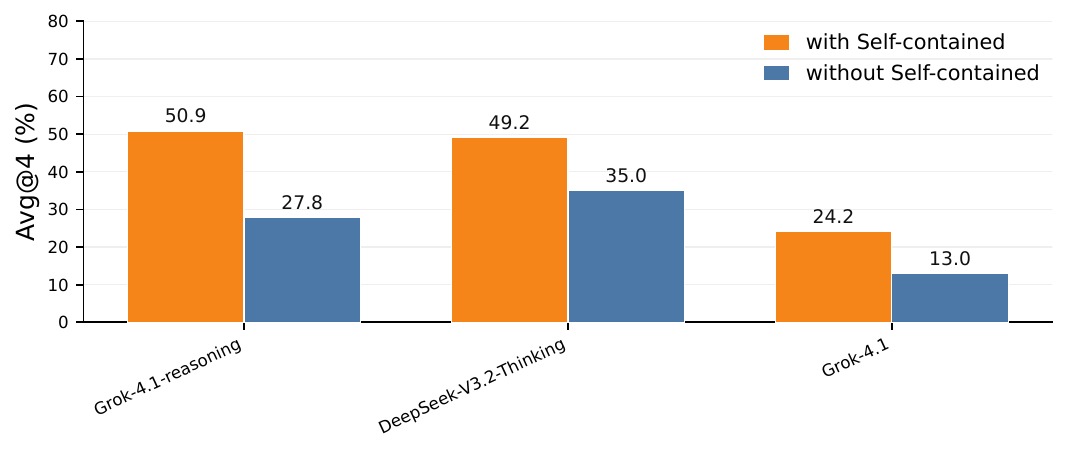}
    \caption{\textbf{Self-containedness ablation.} Removing dependency resolution causes 11\%--23\% performance degradation, confirming that context completeness is essential for valid evaluation.}
    \Description[Grouped bar chart comparing performance with and without self-contained recovery]{A grouped bar chart showing Avg@4 for three models with and without self-contained recovery. With self-contained recovery, Grok-4.1-reasoning scores 50.9 percent, DeepSeek-V3.2-Thinking 49.2 percent, and Grok-4.1 24.2 percent. Without self-contained recovery, performance drops to 27.8, 35.0, and 13.0 percent respectively, representing declines of 11 to 23 percentage points.}
    \label{fig:selfcontained_ablation}
\end{figure}

To verify that our self-contained recovery agent effectively addresses the context incompleteness problem, we compare the performance of three models (Grok-4.1-reasoning, DeepSeek-V3.2-Thinking, and Grok-4.1)  on a non–self-contained variant of Mask-ProofBench.
Figure~\ref{fig:selfcontained_ablation} shows consistent performance degradation under the non–self-contained variant of the benchmark.
Model performance drops by 11\% to 23\%, with reasoning-enhanced models exhibiting larger declines, indicating that model performance strongly depends on access to the resolved proof context.
Notably, models still achieve moderate accuracy (13\%–35\%) without context resolution, suggesting that context incompleteness does not affect all masked instances.
This is because even in non–self-contained proofs, not every masked step requires the missing original proof context for correct reconstruction.
However, the non-negligible performance loss confirms that the self-containedness of research-level mathematical proofs affects a substantial portion of the benchmark, and that our dependency resolution effectively mitigates this issue.

\subsubsection{Ablation Study: Effectiveness of Our Agentic-Masking}
\label{sec:antihack}

A valid instance of mask-in-the-context problem must resist shortcut solutions. If models can achieve high scores by exploiting superficial cues (e.g. positional patterns, local context copying), the evaluation fails to measure genuine reasoning. We validate our mask instances by comparing against random step selection.

\paragraph{Agentic Masking Enables Reliable Model Differentiation.}
To evaluate the effectiveness of our agentic masking strategy, we introduce a random-masking baseline.
The only difference between random masking baseline and Mask-ProofBench is that the former selects mask steps uniformly at random, without any position-selection preferences.
By comparing LLM performance on Mask-ProofBench and this baseline, we show that our mask selection approach leads to more reliable differentiation of models based on their reasoning capability.

Figure~\ref{fig:distortion_scatter} shows that random masking substantially inflates Avg@4 across models. For instance, DeepSeek-V3.2 rises from 29.0\% to 74.4\% (+45.4\%), and Qwen3-14B from 8.1\% to 50.6\% (+42.5\%). Crucially, the score inflation caused by random masking is highly asymmetric. standard models gain disproportionately more, leading to a strong compression of the reasoning advantage observed under agentic masking. In the DeepSeek family, agentic masking yields a clear performance separation, with DeepSeek-V3.2-Thinking achieving 49.2\% and DeepSeek-V3.2 achieving 29.0\%. Under random masking, both models reach nearly identical performance levels (74.0\% vs.\ 74.4\%), eliminating the gap observed under agentic masking. 
This suggests that a large fraction of randomly masked steps can be completed without true multi-step reasoning, relying instead on superficial pattern cues.
Furthermore, random masking leads to unstable and misleading model rankings, where weaker standard models can match or even surpass stronger reasoning-enhanced models, contradicting the expected ranking based on model scale and reasoning capability.

\paragraph{Truncate ablation.}
To address concerns that later proof lines may reveal the masked step, we conduct a separate leakage audit on a fixed diagnostic subset. For each example, we construct a truncated variant by removing the suffix context after the mask, while retaining one complete following mathematical expression to preserve local proof coherence. On this leakage-audit subset, Qwen3-30B-A3B-Instruct changes from 23.79\% to 23.14\% Avg@4, and Qwen3-30B-A3B-Thinking changes from 24.35\% to 20.79\% Avg@4. The performance does not collapse under suffix truncation, suggesting that suffix context is not the dominant source of model performance. We interpret this result conservatively: the audit mitigates, but does not eliminate, the possibility of residual leakage in individual cases.

\paragraph{Expert validation of mask selection.}
To validate that selected masks correspond to meaningful inference-critical steps, we collected 63 domain-expert annotations across eight mathematical areas. Experts assessed formula-level completeness, global reasoning dependency, reasoning necessity, and resistance to shortcut completion. The mean rubric scores were consistently high across algebraic geometry, algebraic topology, PDE/analysis, classical analysis and ODEs, combinatorics, differential geometry, numerical analysis, and optimization/control, supporting that the selected masks are generally valid reasoning targets rather than superficial infilling tasks. The detailed breakdown is provided in Appendix~\ref{app:expert_validation}.

\begin{figure*}[t]
\centering
\includegraphics[width=\textwidth]{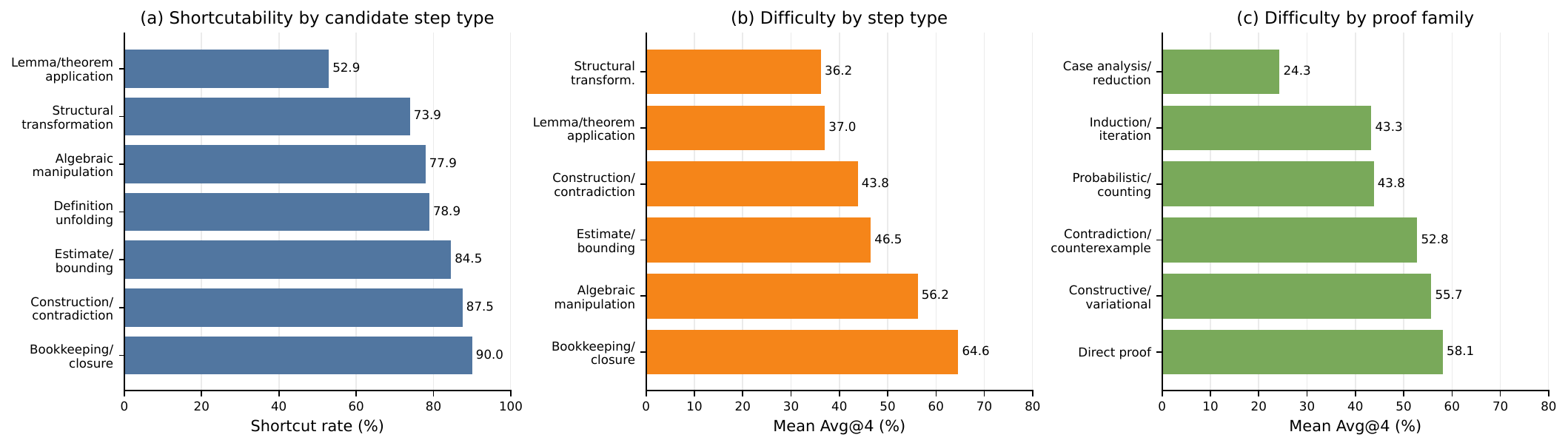}
\caption{Shortcutability and difficulty breakdowns. (a)~Shortcut rates across step types, measured on 30 expert-selected proofs; higher values indicate more shortcut-solvable candidates. (b),~(c)~Mean Avg@4 on Mask-ProofBench by step type and proof family; lower values indicate harder reconstruction.}
\label{fig:compact_breakdowns}
\Description[Three-panel breakdown figure]{The first panel shows shortcut rates by candidate category, with lemma or theorem application the lowest at 52.9 percent and bookkeeping or closure the highest at 90.0 percent. The second panel shows mean Avg@4 by step type, ranging from 36.2 percent for structural transformation to 64.6 percent for bookkeeping or closure. The third panel shows mean Avg@4 by proof family, ranging from 24.3 percent for case analysis or reduction to 58.1 percent for direct proof.}
\end{figure*}

\paragraph{Shortcutability and difficulty analysis.} To clarify which masked steps are genuinely challenging versus easily predictable, and to characterize the benchmark's effective difficulty distribution, we conducted two analyses. First, we masked every candidate position in 30 expert-selected representative proofs and analyzed the shortcut rate of each step type by monitoring LLMs' solving processes alongside small-model performance. Figure~\ref{fig:compact_breakdowns}(a) reports these rates: as hypothesized, algebraic manipulations are significantly more shortcut-prone than lemma/theorem applications (77.9\% vs.\ 52.9\%), and bookkeeping/closure reaches 90.0\%. This asymmetry explains why random masking disproportionately inflates standard-model scores in Figure~\ref{fig:distortion_scatter}. Second, we evaluated three representative models (GPT-5.2 High, Gemini-3-Flash-preview, and Grok-4.1-reasoning) on Mask-ProofBench and report the mean Avg@4 averaged across the three models. Figure~\ref{fig:compact_breakdowns}(b) and~(c) show that structural transformations and lemma/theorem applications are the hardest step types (mean Avg@4 of 36.2\% and 37.0\%), and case analysis/reduction is the hardest proof family (24.3\%), compared to direct proof at 58.1\%.

\begin{table*}[!ht] 
\centering
\small
\caption{Source-level breakdown of our Mask-Proof (Avg@4, \%) compared with the official expert evaluation (\% of total points) on IMO-ProofBench-Advanced.}
\label{tab:mask_proxy_breakdown_filtered}
\setlength{\tabcolsep}{6pt}
\begin{tabular}{l cccc cccc}
\toprule
\multirow{2}{*}{\textbf{Model}} & \multicolumn{4}{c}{\textbf{Mask-Proof (this work): Avg@4 (\%)}} & \multicolumn{4}{c}{\textbf{Expert (official): score (\%)}} \\
\cmidrule(lr){2-5} \cmidrule(lr){6-9}
& Adv & Novel & IMO2024 & USAMO2025 & Adv & Novel & IMO2024 & USAMO2025 \\
\midrule
Number of Problems & 35 & 22 & 8 & 5 & 30 & 18 & 6 & 6 \\
\midrule
o3 & 52.9 & 44.3 & 53.1 & \textbf{90.0} & \textbf{20.5} & 15.1 & 4.8 & \textbf{52.4} \\
Gemini 2.5 Pro & \textbf{58.6} & \textbf{55.7} & 46.9 & \textbf{90.0} & 17.6 & \textbf{15.9} & \textbf{7.1} & 33.3 \\
Grok 4.1 Reasoning & 50.0 & 43.2 & \textbf{53.1} & 75.0 & 18.6 & 19.8 & 16.7 & 16.7 \\
o4-mini & 42.1 & 33.0 & 46.9 & 75.0 & 11.4 & 8.7 & 7.1 & 23.8 \\
Kimi-K2 & 30.7 & 23.9 & 25.0 & 70.0 & 7.1 & 4.0 & 2.4 & 21.4 \\
Qwen3-235B & 34.3 & 34.1 & 31.2 & 40.0 & 5.2 & 7.1 & 0.0 & 4.8 \\
DeepSeek V3 & 32.9 & 30.7 & 25.0 & 55.0 & 4.3 & 6.3 & 2.4 & 0.0 \\
\midrule
Spearman $\rho$ (Expert vs. Mask) & 0.79 & 0.86 & 0.73 & 0.89 & \multicolumn{4}{c}{--} \\
\bottomrule
\end{tabular}
\end{table*}

\subsection{Reproducibility of the Mask-Proof Pipeline (RQ2)}

\subsubsection{Self-Contained Recovery Reproducibility}
\label{sec:selfcontained-reproducibility}
To evaluate the reproducibility of our self-contained module, we run the pipeline twice independently on 20 sampled proofs from Mask-ProofBench. Three of these proofs are intrinsically self-contained and require no additional context.

\textbf{Explicit dependencies.} For references with explicit labels (via \texttt{\textbackslash ref}, \texttt{\textbackslash eqref}), we directly compare the resolved label sets. The two runs achieve 97.2\% recall: out of 36 unique labels across both runs, 35 appear in both. The single discrepancy is an optional theorem reference included by one run but not the other.
The high BLEU scores indicate that independently recovered Additional Information blocks are nearly identical across runs, suggesting that the implicit dependency resolution process is stable rather than sensitive to stochastic variation.
\textbf{Implicit dependencies.} For implicit dependencies—such as symbols with closed-form definitions elsewhere—no ground-truth labels exist for direct comparison. We use textual similarity as a proxy: if two runs resolve the same implicit dependencies, they should produce similar Additional Information blocks. The symmetric BLEU-4 achieves a mean of 0.900 and median of 0.974, with all 20 proofs exceeding the 0.3 threshold.

\subsubsection{Mask Selection Reproducibility}

To evaluate the reproducibility of mask selection, we run the agentic masking module twice independently on the same 20 proofs used in Section~\ref{sec:selfcontained-reproducibility}. Both runs produce 28 masks in total. We measure reproducibility by checking how many masks from the first run appear in the second.

After normalizing each masked formula—stripping outer math delimiters (e.g., \texttt{\$...\$}, \texttt{\textbackslash[...\textbackslash]}) and whitespace—we check for exact matches within the same proof: 19 out of 28 masks (67.9\%) are reproduced. We manually annotate all 9 non-reproduced masks into three categories:
\begin{itemize}
    \item \textbf{Semantically proximate (6/9):} The two runs select different points along the same formula chain, e.g., an intermediate equation vs.\ the final result.
    \item \textbf{Same proof, different sub-step (1/9):} Both masks address related reasoning but test different skills, e.g., an error bound vs.\ the parameter choice that optimizes it.
    \item \textbf{Genuinely different (2/9):} The masks target unrelated mathematical content.
\end{itemize}
Counting semantically proximate cases as successful reproduction, the consistency rate rises from 67.9\% to 89.3\%. Only 2 out of 28 masks (7.1\%) target genuinely different content.

Taken together, these results answer RQ2 by showing that the Mask-Proof pipeline is reproducible at both the self-contained recovery and mask selection stages. 
Independent runs produce highly consistent self-contained contexts and largely overlapping masks, with most discrepancies reflecting minor semantic variations rather than genuinely different problem constructions. For released runs, we record the Codex CLI version, model setting, prompt, input hash, and deterministic post-processing scripts so that LLM-assisted stages can be rerun and audited even though they are not bitwise deterministic.

\subsection{Consistency Between Human Experts and Mask-ProofJudge (RQ3)}
\label{sec:judge_reliability}

Since our benchmark masks pure LaTeX mathematical formulas---where a single ground truth may have multiple semantically equivalent representations (e.g., notation variants, unsimplified terms)---we validate that Mask-ProofJudge can reliably assess equivalence.

We constructed a validation set of 1273 samples randomly drawn from actual responses of four different models(Gemini-3-Flash-preview, DeepSeek-V3.2-Thinking, Kimi-K2, and Grok-4.1), containing a balanced mix of correct answers and plausible but incorrect formulas. Two domain experts independently labeled the set by checking mathematical equivalence against the ground truth. Comparing Mask-ProofJudge's majority-vote verdicts to expert consensus, GPT-OSS-120B(High) achieves \textbf{96.8\% accuracy}, indicating that it reliably distinguishes benign notational variation from substantive mathematical errors.

\subsection{Cross-Dataset Generalization Ability (RQ4)}
To evaluate whether the Mask-Proof pipeline generalizes beyond our curated benchmark, we apply it to an external dataset with existing expert evaluations, enabling direct comparison between our automated mask-based scoring and human expert judgments.

We apply the Mask-Proof pipeline to all 30 PB-Advanced proofs from IMO-ProofBench~\cite{luong-etal-2025-towards}, producing 35 masks across three sources: Novel (22), IMO2024 (8), and USAMO2025 (5). We evaluate seven models using both our mask-based Avg@4 metric and the official expert scores. Table~\ref{tab:mask_proxy_breakdown_filtered} presents the source-level breakdown and ranking consistency measured by Spearman's $\rho$.

The mask-based evaluation achieves Spearman $\rho = 0.79$ overall, reaching $0.89$ on USAMO2025 and $0.86$ on Novel. This correlation is obtained entirely automatically—no human grading, no problem-specific rubrics, and no per-task scoring guidelines—whereas IMO-ProofBench~\cite{luong-etal-2025-towards} expert evaluation requires domain experts to design rubrics and manually grade every response. Moreover, Mask-ProofJudge uses GPT-OSS-120B High as an open-source judge, rather than relying on frontier proprietary models or per-response human grading. Despite this modest resource requirement, the resulting model rankings closely track expert consensus, demonstrating that Mask-Proof's discriminative power comes primarily from agentic masking of key formula steps rather than expensive evaluation infrastructure.

The moderate correlation on IMO2024 ($\rho = 0.73$) reflects a granularity mismatch: expert evaluations consider holistic proof structure and award partial credit for incomplete solutions, while mask-based evaluation targets localized step-level reasoning. These two paradigms are complementary—Mask-Proof excels at scalable, reproducible assessment of specific reasoning steps, while expert evaluation captures broader proof quality dimensions. Overall, Mask-Proof generalizes effectively to external proof sources and produces model rankings consistent with expert consensus, positioning it as a practical, scalable alternative for evaluating LLMs on research-level mathematical reasoning.

\section{Conclusion}
\label{sec:conclusion}
In this work, we presented \textbf{Mask-Proof}, an automatic pipeline for generating refreshable masked proof tasks and evaluating large language models on mathematical proofs. Mask-Proof generated
\textbf{Mask-ProofBench}, a curated benchmark of 292 masked proof-step instances spanning 35 mathematical fields. Extensive experiments show that \textbf{Mask-Proof} can remove shortcut-solvable instances, and evaluations of \textbf{Mask-ProofBench} and Masked IMO-ProofBench reveal its effectiveness, reproducibility, and generalization across diverse proof sources.

\section{Limitations and Ethical Considerations}

\textbf{Limitations.}
Mask-Proof currently relies on raw LaTeX sources; extending coverage to PDFs, scanned documents, and web-native mathematical text would require reliable mathematical OCR or multimodal parsing. The LLM-assisted components for dependency recovery and mask selection have been validated through reproducibility analyses and expert review, though the process is not bitwise deterministic, and our leakage and shortcutability analyses mitigate but do not eliminate the possibility of residual cues. The LLM-based judge (96.8\% expert agreement) may also inherit biases from its underlying model.

\textbf{Ethical Considerations.}
All source papers are drawn from arXiv under licenses permitting
redistribution for research; we release only extracted proof fragments
with attribution. To mitigate benchmark contamination, we restrict our
corpus to the most recent three months of submissions and recommend
periodic refresh. This work involves no human subjects.

\begin{acks}
We thank the anonymous reviewers for their constructive 
feedback. This work was supported by the ``111 Center'' 
(No.\ B26023).
\end{acks}

\bibliographystyle{ACM-Reference-Format}
\bibliography{references}

@misc{bubeck2025earlyscienceaccelerationexperiments,
  title = {Early science acceleration experiments with GPT-5},
  author = {Bubeck, Sébastien and Coester, Christian and Eldan, Ronen and Gowers, Timothy and Lee, Yin Tat and Lupsasca, Alexandru and Sawhney, Mehtaab and Scherrer, Robert and Sellke, Mark and Spears, Brian K. and Unutmaz, Derya and Weil, Kevin and Yin, Steven and Zhivotovskiy, Nikita},
  year = {2025},
  eprint = {2511.16072},
  archivePrefix = {arXiv},
  primaryClass = {cs.CL},
  url = {https://arxiv.org/abs/2511.16072}
}

@inproceedings{NEURIPS2022_9d560961,
  title = {Chain-of-Thought Prompting Elicits Reasoning in Large Language Models},
  author = {Wei, Jason and Wang, Xuezhi and Schuurmans, Dale and Bosma, Maarten and ichter, brian and Xia, Fei and Chi, Ed and Le, Quoc V and Zhou, Denny},
  booktitle = {Advances in Neural Information Processing Systems},
  volume = {35},
  pages = {24824--24837},
  year = {2022},
  url = {https://papers.nips.cc/paper_files/paper/2022/hash/9d5609613524ecf4f15af0f7b31abca4-Abstract-Conference.html}
}

@article{trinh2024solving,
  title = {Solving olympiad geometry without human demonstrations},
  author = {Trinh, Trieu H. and Wu, Yuhuai and Le, Quoc V. and He, He and Luong, Thang},
  journal = {Nature},
  volume = {625},
  number = {7995},
  pages = {476--482},
  year = {2024},
  doi = {10.1038/s41586-023-06747-5}
}

@misc{cobbe2021gsm8k,
  title = {Training Verifiers to Solve Math Word Problems},
  author = {Cobbe, Karl and Kosaraju, Vineet and Bavarian, Mohammad and Chen, Mark and Jun, Heewoo and Kaiser, Lukasz and Plappert, Matthias and Tworek, Jerry and Hilton, Jacob and Nakano, Reiichiro and Hesse, Christopher and Schulman, John},
  year = {2021},
  eprint = {2110.14168},
  archivePrefix = {arXiv},
  primaryClass = {cs.LG},
  url = {https://arxiv.org/abs/2110.14168}
}

@article{Geirhos_2020,
  title = {Shortcut learning in deep neural networks},
  author = {Geirhos, Robert and Jacobsen, Jörn-Henrik and Michaelis, Claudio and Zemel, Richard and Brendel, Wieland and Bethge, Matthias and Wichmann, Felix A.},
  journal = {Nature Machine Intelligence},
  volume = {2},
  number = {11},
  pages = {665--673},
  year = {2020},
  doi = {10.1038/s42256-020-00257-z}
}

@misc{pandit2025hard2verifysteplevelverificationbenchmark,
  title = {Hard2Verify: A Step-Level Verification Benchmark for Open-Ended Frontier Math},
  author = {Pandit, Shrey and Xu, Austin and Nguyen, Xuan-Phi and Ming, Yifei and Xiong, Caiming and Joty, Shafiq},
  year = {2025},
  eprint = {2510.13744},
  archivePrefix = {arXiv},
  primaryClass = {cs.AI},
  url = {https://arxiv.org/abs/2510.13744}
}

@inproceedings{zhengJudgingLLMasJudge,
  title = {Judging {LLM}-as-a-Judge with {MT}-Bench and Chatbot Arena},
  author = {Zheng, Lianmin and Chiang, Wei-Lin and Sheng, Ying and Zhuang, Siyuan and Wu, Zhanghao and Zhuang, Yonghao and Lin, Zi and Li, Zhuohan and Li, Dacheng and Xing, Eric P. and Zhang, Hao and Gonzalez, Joseph E. and Stoica, Ion},
  booktitle = {Advances in Neural Information Processing Systems},
  volume = {36},
  pages = {46595--46623},
  year = {2023},
  url = {https://papers.nips.cc/paper_files/paper/2023/hash/91f18a1287b398d378ef22505bf41832-Abstract-Datasets_and_Benchmarks.html}
}

@misc{shao2025deepseekmathv2selfverifiablemathematicalreasoning,
  title = {{DeepSeekMath}-{V2}: Towards Self-Verifiable Mathematical Reasoning},
  author = {Shao, Zhihong and Luo, Yuxiang and Lu, Chengda and Ren, Z. Z. and Hu, Jiewen and Ye, Tian and Gou, Zhibin and Ma, Shirong and Zhang, Xiaokang},
  year = {2025},
  eprint = {2511.22570},
  archivePrefix = {arXiv},
  primaryClass = {cs.AI},
  url = {https://arxiv.org/abs/2511.22570}
}

@inproceedings{liu-etal-2025-storm,
  title = {{STORM}-{BORN}: A Challenging Mathematical Derivations Dataset Curated via a Human-in-the-Loop Multi-Agent Framework},
  author = {Liu, Wenhao and Lu, Zhenyi and Hu, Xinyu and Zhang, Jerry and Li, Dailin and Cen, Jiacheng and Cao, Huilin and Wang, Haiteng and Li, Yuhan and Kun, Xie and Li, Dandan and Zhang, Pei and Zhang, Chengbo and Ren, Yuxiang and Huang, Xiaohong and Ma, Yan},
  booktitle = {Findings of the Association for Computational Linguistics: ACL 2025},
  pages = {23938--23958},
  year = {2025},
  doi = {10.18653/v1/2025.findings-acl.1227}
}

@inproceedings{hendrycksmath2021,
  title = {Measuring Mathematical Problem Solving With the {MATH} Dataset},
  author = {Hendrycks, Dan and Burns, Collin and Kadavath, Saurav and Arora, Akul and Basart, Steven and Tang, Eric and Song, Dawn and Steinhardt, Jacob},
  booktitle = {Proceedings of the Neural Information Processing Systems Track on Datasets and Benchmarks},
  year = {2021},
  url = {https://arxiv.org/abs/2103.03874}
}

@inproceedings{gao2024omnimathuniversalolympiadlevel,
  title = {{Omni-MATH}: A Universal Olympiad Level Mathematic Benchmark for Large Language Models},
  author = {Gao, Bofei and Song, Feifan and Yang, Zhe and Cai, Zefan and Miao, Yibo and Dong, Qingxiu and Li, Lei and Ma, Chenghao and Chen, Liang and Xu, Runxin and Tang, Zhengyang and Wang, Benyou and Zan, Daoguang and Quan, Shanghaoran and Zhang, Ge and Sha, Lei and Zhang, Yichang and Ren, Xuancheng and Liu, Tianyu and Chang, Baobao},
  booktitle = {International Conference on Learning Representations},
  year = {2025},
  url = {https://openreview.net/forum?id=yaqPf0KAlN}
}

@article{shankar2025counting,
  title = {Counting integral points on symmetric varieties with applications to arithmetic statistics},
  author = {Shankar, Arul and Siad, Artane and Swaminathan, Ashvin A.},
  journal = {Proceedings of the London Mathematical Society},
  volume = {130},
  number = {4},
  pages = {e70039},
  year = {2025},
  doi = {10.1112/plms.70039}
}

@inproceedings{tyen-etal-2024-llms,
  title = {{LLM}s cannot find reasoning errors, but can correct them given the error location},
  author = {Tyen, Gladys and Mansoor, Hassan and Carbune, Victor and Chen, Peter and Mak, Tony},
  booktitle = {Findings of the Association for Computational Linguistics: ACL 2024},
  pages = {13894--13908},
  year = {2024},
  doi = {10.18653/v1/2024.findings-acl.826}
}

@inproceedings{luong-etal-2025-towards,
  title = {Towards Robust Mathematical Reasoning},
  author = {Luong, Thang and Hwang, Dawsen and Nguyen, Hoang H. and Ghiasi, Golnaz and Chervonyi, Yuri and Seo, Insuk and Kim, Junsu and Bingham, Garrett and Lee, Jonathan and Mishra, Swaroop and Zhai, Alex and Hu, Huiyi and Michalewski, Henryk and Kim, Jimin and Ahn, Jeonghyun and Bae, Junhwi and Song, Xingyou and Trinh, Trieu Hoang and Le, Quoc V. and Jung, Junehyuk},
  booktitle = {Proceedings of the 2025 Conference on Empirical Methods in Natural Language Processing},
  pages = {35418--35442},
  year = {2025},
  doi = {10.18653/v1/2025.emnlp-main.1794}
}

@inproceedings{chen2025reform,
  title     = {{ReForm}: Reflective Autoformalization with Prospective Bounded Sequence Optimization},
  author    = {Chen, Guoxin and Wu, Jing and Chen, Xinjie and Zhao, Wayne Xin and Song, Ruihua and Li, Chengxi and Fan, Kai and Liu, Dayiheng and Liao, Minpeng},
  booktitle = {International Conference on Learning Representations},
  year      = {2026},
  url       = {https://openreview.net/forum?id=KfxRzCmRSX}
}

@inproceedings{zhang2025realmath,
  title = {{RealMath}: A Continuous Benchmark for Evaluating Language Models on Research-Level Mathematics},
  author = {Zhang, Jie and Petrui, Cezara and Nikolić, Kristina and Tramèr, Florian},
  booktitle = {NeurIPS 2025 Track on Datasets and Benchmarks},
  year = {2025},
  url = {https://openreview.net/forum?id=RBssYVpQEr}
}

@inproceedings{lightman2023verify,
  title = {Let's Verify Step by Step},
  author = {Lightman, Hunter and Kosaraju, Vineet and Burda, Yuri and Edwards, Harrison and Baker, Bowen and Lee, Teddy and Leike, Jan and Schulman, John and Sutskever, Ilya and Cobbe, Karl},
  booktitle = {International Conference on Learning Representations},
  year = {2024},
  url = {https://openreview.net/forum?id=v8L0pN6EOi}
}

@inproceedings{zheng2021minif2f,
  title = {{miniF2F}: a cross-system benchmark for formal Olympiad-level mathematics},
  author = {Zheng, Kunhao and Han, Jesse Michael and Polu, Stanislas},
  booktitle = {International Conference on Learning Representations},
  year = {2022},
  url = {https://openreview.net/forum?id=9ZPegFuFTFv}
}

@inproceedings{he2024olympiadbench,
  title = {{O}lympiad{B}ench: A Challenging Benchmark for Promoting {AGI} with Olympiad-Level Bilingual Multimodal Scientific Problems},
  author = {He, Chaoqun and Luo, Renjie and Bai, Yuzhuo and Hu, Shengding and Thai, Zhen and Shen, Junhao and Hu, Jinyi and Han, Xu and Huang, Yujie and Zhang, Yuxiang and Liu, Jie and Qi, Lei and Liu, Zhiyuan and Sun, Maosong},
  booktitle = {Proceedings of the 62nd Annual Meeting of the Association for Computational Linguistics (Volume 1: Long Papers)},
  pages = {3828--3850},
  year = {2024},
  doi = {10.18653/v1/2024.acl-long.211}
}

@inproceedings{tsoukalas2024putnambench,
  title = {{PutnamBench}: Evaluating Neural Theorem-Provers on the Putnam Mathematical Competition},
  author = {Tsoukalas, George and Lee, Jasper and Jennings, John and Xin, Jimmy and Ding, Michelle and Jennings, Michael and Thakur, Amitayush and Chaudhuri, Swarat},
  booktitle = {NeurIPS 2024 Track on Datasets and Benchmarks},
  year = {2024},
  url = {https://openreview.net/forum?id=ChKCF75Ocd}
}

@misc{luo2024automatedprocesssupervision,
  title = {Improve Mathematical Reasoning in Language Models by Automated Process Supervision},
  author = {Luo, Liangchen and Liu, Yinxiao and Liu, Rosanne and Phatale, Samrat and Guo, Meiqi and Lara, Harsh and Li, Yunxuan and Shu, Lei and Zhu, Yun and Meng, Lei and Sun, Jiao and Rastogi, Abhinav},
  year = {2024},
  eprint = {2406.06592},
  archivePrefix = {arXiv},
  primaryClass = {cs.CL},
  url = {https://arxiv.org/abs/2406.06592}
}

@inproceedings{zheng-etal-2025-processbench,
  title = {{P}rocess{B}ench: Identifying Process Errors in Mathematical Reasoning},
  author = {Zheng, Chujie and Zhang, Zhenru and Zhang, Beichen and Lin, Runji and Lu, Keming and Yu, Bowen and Liu, Dayiheng and Zhou, Jingren and Lin, Junyang},
  booktitle = {Proceedings of the 63rd Annual Meeting of the Association for Computational Linguistics (Volume 1: Long Papers)},
  pages = {1009--1024},
  year = {2025},
  doi = {10.18653/v1/2025.acl-long.50}
}

@misc{glazer2024frontiermath,
  title = {{FrontierMath}: A Benchmark for Evaluating Advanced Mathematical Reasoning in {AI}},
  author = {Glazer, Elliot and Erdil, Ege and Besiroglu, Tamay and Chicharro, Diego and Chen, Evan and Gunning, Alex and Olsson, Caroline Falkman and Denain, Jean-Stanislas and Ho, Anson and de Oliveira Santos, Emily and Järviniemi, Olli and Barnett, Matthew and Sandler, Robert and Vrzala, Matej and Sevilla, Jaime and Ren, Qiuyu and Pratt, Elizabeth and Levine, Lionel and Barkley, Grant and Stewart, Natalie and Grechuk, Bogdan and Grechuk, Tetiana and Enugandla, Shreepranav Varma and Wildon, Mark},
  year = {2024},
  eprint = {2411.04872},
  archivePrefix = {arXiv},
  primaryClass = {cs.AI},
  url = {https://arxiv.org/abs/2411.04872}
}

@inproceedings{huang2024selfcorrect,
  title = {Large Language Models Cannot Self-Correct Reasoning Yet},
  author = {Huang, Jie and Chen, Xinyun and Mishra, Swaroop and Zheng, Huaixiu Steven and Yu, Adams Wei and Song, Xinying and Zhou, Denny},
  booktitle = {International Conference on Learning Representations},
  year = {2024},
  url = {https://openreview.net/forum?id=IkmD3fKBPQ}
}

@misc{schmitt2025improofbenchbenchmarkingairesearchlevel,
  title = {{IMProofBench}: Benchmarking {AI} on Research-Level Mathematical Proof Generation},
  author = {Schmitt, Johannes and Bérczi, Gergely and Dekoninck, Jasper and Feusi, Jeremy and Gehrunger, Tim and Appenzeller, Raphael and Bryan, Jim and Canova, Niklas and de Wolff, Timo and Gaia, Filippo and van Garrel, Michel and Hashemi, Baran and Holmes, David and Iribar Lopez, Aitor and Jaeck, Victor and Jørgensen, Martina and Kelk, Steven and Kuhlmann, Stefan and Kurpisz, Adam and Meroni, Chiara and Metzler, Ingmar and Möller, Martin and Muñoz-Echániz, Samuel and Nowak, Robert and Oberdieck, Georg and Platt, Daniel and Possamaï, Dylan and Ribeiro, Gabriel and Sánchez Galán, Raúl and Sun, Zheming and Teichmann, Josef and Thomas, Richard P. and Vial, Charles},
  year = {2025},
  eprint = {2509.26076},
  archivePrefix = {arXiv},
  primaryClass = {cs.CL},
  url = {https://arxiv.org/abs/2509.26076}
}

@misc{chen2022program,
  title = {Program of Thoughts Prompting: Disentangling Computation from Reasoning for Numerical Reasoning Tasks},
  author = {Chen, Wenhu and Ma, Xueguang and Wang, Xinyi and Cohen, William W.},
  year = {2022},
  eprint = {2211.12588},
  archivePrefix = {arXiv},
  primaryClass = {cs.CL},
  url = {https://arxiv.org/abs/2211.12588}
}

@misc{ma2025reliablefinegrainedevaluationnatural,
  title = {Reliable Fine-Grained Evaluation of Natural Language Math Proofs},
  author = {Ma, Wenjie and Cojocaru, Andrei and Kolhe, Neel and Louie, Bradley and Sharif, Robin Said and Zhang, Haihan and Zhuang, Vincent and Zaharia, Matei and Min, Sewon},
  year = {2025},
  eprint = {2510.13888},
  archivePrefix = {arXiv},
  primaryClass = {cs.CL},
  url = {https://arxiv.org/abs/2510.13888}
}

@misc{yang2025utmathmathevaluationunit,
  title = {{UTMath}: Math Evaluation with Unit Test via Reasoning-to-Coding Thoughts},
  author = {Yang, Bo and Yang, Qingping and Ma, Yingwei and Liu, Runtao},
  year = {2025},
  eprint = {2411.07240},
  archivePrefix = {arXiv},
  primaryClass = {cs.CL},
  url = {https://arxiv.org/abs/2411.07240}
}

@inproceedings{zhu2023dyval,
  title = {{DyVal}: Dynamic Evaluation of Large Language Models for Reasoning Tasks},
  author = {Zhu, Kaijie and Chen, Jiaao and Wang, Jindong and Gong, Neil Zhenqiang and Yang, Diyi and Xie, Xing},
  booktitle = {International Conference on Learning Representations},
  year = {2024},
  url = {https://openreview.net/forum?id=gjfOL9z5Xr}
}

@misc{yu2023metamath,
  title = {{MetaMath}: Bootstrap Your Own Mathematical Questions for Large Language Models},
  author = {Yu, Longhui and Jiang, Weisen and Shi, Han and Yu, Jincheng and Liu, Zhengying and Zhang, Yu and Kwok, James T. and Li, Zhenguo and Weller, Adrian and Liu, Weiyang},
  year = {2023},
  eprint = {2309.12284},
  archivePrefix = {arXiv},
  primaryClass = {cs.CL},
  url = {https://arxiv.org/abs/2309.12284}
}

@inproceedings{kim2023prometheus,
  title = {{Prometheus}: Inducing Fine-Grained Evaluation Capability in Language Models},
  author = {Kim, Seungone and Shin, Jamin and Cho, Yejin and Jang, Joel and Longpre, Shayne and Lee, Hwaran and Yun, Sangdoo and Shin, Seongjin and Kim, Sungdong and Thorne, James and Seo, Minjoon},
  booktitle = {International Conference on Learning Representations},
  year = {2024},
  url = {https://openreview.net/forum?id=8euJaTveKw}
}

@misc{thakur2025judgingjudgesevaluatingalignment,
  title = {Judging the Judges: Evaluating Alignment and Vulnerabilities in {LLM}s-as-Judges},
  author = {Thakur, Aman Singh and Choudhary, Kartik and Ramayapally, Venkat Srinik and Vaidyanathan, Sankaran and Hupkes, Dieuwke},
  year = {2025},
  eprint = {2406.12624},
  archivePrefix = {arXiv},
  primaryClass = {cs.CL},
  url = {https://arxiv.org/abs/2406.12624}
}

@inproceedings{ke2024critiquellminformativecritiquegeneration,
  title = {{CritiqueLLM}: Towards an Informative Critique Generation Model for Evaluation of Large Language Model Generation},
  author = {Ke, Pei and Wen, Bosi and Feng, Andrew and Liu, Xiao and Lei, Xuanyu and Cheng, Jiale and Wang, Shengyuan and Zeng, Aohan and Dong, Yuxiao and Wang, Hongning and Tang, Jie and Huang, Minlie},
  booktitle = {Proceedings of the 62nd Annual Meeting of the Association for Computational Linguistics (Volume 1: Long Papers)},
  pages = {13034--13054},
  year = {2024},
  doi = {10.18653/v1/2024.acl-long.704}
}

@misc{balunović2026matharenaevaluatingllmsuncontaminated,
  title = {{MathArena}: Evaluating {LLM}s on Uncontaminated Math Competitions},
  author = {Balunović, Mislav and Dekoninck, Jasper and Petrov, Ivo and Jovanović, Nikola and Vechev, Martin},
  year = {2026},
  eprint = {2505.23281},
  archivePrefix = {arXiv},
  primaryClass = {cs.AI},
  url = {https://arxiv.org/abs/2505.23281}
}

\appendix

\onecolumn

\section{Mask Rubrics.}
\label{app:maskrubric}
Our mask rubrics are as follows:
\begin{itemize}
\item \textbf{Formula-level completeness.}
Each mask must correspond to a \emph{complete mathematical expression or formula}, rather than a fragment of an equation or a syntactic sub-component. Partial masking of symbols, coefficients, or operators is disallowed.

\item \textbf{Global reasoning dependency.}
A valid mask should not be solvable by inspecting only its immediate local context. Correct recovery must require integrating information from the theorem statement, earlier definitions or lemmas, and the global logical structure of the proof.

\item \textbf{Reasoning density and necessity.}
Masked steps should represent logically critical steps (e.g., non-trivial formulas, structural identities, or key transformations) whose absence would break the proof. Trivial algebraic rearrangements or routine substitutions are excluded.

\item \textbf{Non-templated and non-heuristic.}
Masks must not be recoverable via pattern matching, standard proof templates, symmetry guessing, backward inference from later lines, or memorized formula instantiation. Any mask that can be solved without genuine mathematical reasoning is considered invalid.

\item \textbf{Quality-controlled multiplicity.}
For each proof, we allow between $0$ and $3$ masks, determined by the proof’s conceptual depth, technical sophistication, and reasoning density. Proofs that are low-quality, overly short, informal, or mathematically trivial may yield zero masks.
\end{itemize}

\section{Additional Validation Details}
\label{app:expert_validation}

\paragraph{Expert validation of mask positions.}
Table~\ref{tab:expert_validation} summarizes the 63 domain-expert annotations used to validate whether agentically selected masks are meaningful reasoning targets rather than superficial completion tasks.

\begin{table}[H]
\centering
\caption{Expert validation of mask positions across mathematical areas.}
\label{tab:expert_validation}
\small
\begin{tabular}{lccc}
\toprule
Area & \#Ann. & Mean Score & High Quality \\
\midrule
Algebraic Geometry & 9 & 8.22 & 9/9 \\
Algebraic Topology & 7 & 8.29 & 7/7 \\
PDE / Analysis & 8 & 7.50 & 6/8 \\
Classical Analysis and ODEs & 10 & 8.00 & 9/10 \\
Combinatorics & 8 & 7.00 & 6/8 \\
Differential Geometry & 8 & 8.00 & 6/8 \\
Numerical Analysis & 7 & 8.86 & 7/7 \\
Optimization and Control & 6 & 7.67 & 5/6 \\
\bottomrule
\end{tabular}
\end{table}

\paragraph{Truncation leakage audit.}
Table~\ref{tab:leakage_audit} reports the suffix-truncation audit used to test whether later proof context is the dominant source of model performance. These results should be read as evidence that the risk is mitigated, not as a proof that leakage is impossible.

\begin{table}[H]
\centering
\caption{Suffix-truncation leakage audit on the leakage-audit split.}
\label{tab:leakage_audit}
\small
\begin{tabular}{lcc}
\toprule
Model & Original Avg@4 & Truncated Avg@4 \\
\midrule
Qwen3-30B-A3B-Instruct & 23.79 & 23.14 \\
Qwen3-30B-A3B-Thinking & 24.35 & 20.79 \\
\bottomrule
\end{tabular}
\end{table}

\section{Data format of Mask-Proof Samples.}
\label{app:dataformat}


\paragraph{Data format}
We serialize the processed samples in \texttt{jsonl}. Each record contains a unique identifier and the paired question/solution:
\begin{verbatim}
{ 
"index": 6, "arxiv_id": "...", 
"proof_index": 8,
"problem": <<PROBLEM TO PROVE>>, 
"reference_solution": <<PROOF>>,
"mask_text": <<PROOF CONTAINS 1 MASK>>,
"mask_content": <<ORIGINAL TEXT IN THE MASK>>
}
\end{verbatim}
Here \texttt{mask\_text} is the masked prompt (typically formula-only masks), and \texttt{mask\_content} is the corresponding ground-truth proof fragment.


\section{Data Distribution}

\begin{table}[H]
  \caption{Percentage distribution of the top arXiv subject categories of papers used in Mask-ProofBench.}
  \label{tab:datadistributionpreview}
  \begin{center}
        \begin{tabular}{lccc}
        \toprule
        Category  & Percentage  &  Category & Percentage\\
        \midrule
        PDE / Analysis   & 26.47\%  &  Combinatorics  & 2.94\% \\ 
        Mathematical Physics & 5.88\%  &  Numerical Analysis  & 2.94\% \\
        Number Theory  &  5.15\%  &  Optimization and Control  &  2.94\% \\ 
        Differential Geometry & 5.15\% &  Operator Algebras & 2.94\% \\
        Functional Analysis & 4.41\% & Algebraic Geometry & 2.21\% \\
        Representation Theory & 4.41\% & Algebraic Topology & 2.21\% \\
        Probability & 3.68\% & Statistics Theory & 2.21\% \\
        Classical Analysis and ODEs & 2.94\% & Others & 23.52\%\\
        \bottomrule
        \end{tabular}
  \end{center}
\end{table}

\section{Case Study}
\label{app:artanecase}

\textbf{Masked Proof:}
\begin{proof}
We prove the result for $W(K)$; the proof for $W^\vee(K)$ is identical.
Denote the coefficients of $f$ by $a, b, c,$ and $d$, and assume (by replacing
$f$ with a $\mathrm{GL}_2(K)$-translate, if necessary) that $ad \neq 0$.
First, it is easy to check that there exists a distinguished orbit
(resp.\ a $\Delta$-distinguished orbit) in $W(K)$.
The distinguished orbit is represented by the following pair:
\[
(A,B)
=
\left(
\begin{pmatrix}
0 & 0 & \tfrac12 \\
0 & -a & 0 \\
\tfrac12 & 0 & -c
\end{pmatrix},
\begin{pmatrix}
0 & \tfrac12 & 0 \\
\tfrac12 & b & 0 \\
0 & 0 & d
\end{pmatrix}
\right).
\]
The $\Delta$-distinguished orbit with resolvent $f$ is represented by the following pair:
\begin{equation}\label{eq:deltadist}
(A,B)
=
\left(
\begin{pmatrix}
-a & 0 & 0 \\
0 & 0 & 1/2 \\
0 & 1/2 & b/4ad
\end{pmatrix},
\begin{pmatrix}
0 & 0 & 1/2 \\
0 & d & 0 \\
1/2 & 0 & -c/4ad
\end{pmatrix}
\right).
\end{equation}
It only remains to prove uniqueness.
For the distinguished case, let $L$ be the \'etale cubic extension of $K$
corresponding to $f$ under the Delone--Faddeev bijection.
$\mathrm{SL}_3(K)$-orbits on $W(K) \cap \mathrm{Res}^{-1}(f)$ correspond bijectively
to \'etale quartic extensions of $K$ with resolvent $f$ by Bhargava's
parametrization.
An orbit is distinguished if and only if it corresponds to $L \oplus K$,
and so there is a unique distinguished orbit.
We give a much more hands-on proof of uniqueness of the $\Delta$-distinguished orbit.
Let $(A,B)$ (with coefficients $a_{ij}$ and $b_{ij}$) be an element in a
$\Delta$-distinguished orbit with $\mathrm{Res}(A,B)=f$.
By the definition of being $\Delta$-distinguished, there exists a
$\mathbb{P}^1_K \subset \mathbb{P}^2_K$ such that the restriction of $A$ and $B$
to this $\mathbb{P}^1_K$ give reducible binary quadratic forms.
(Note that neither of these binary forms can be $0$, otherwise we would have
$\Delta(A,B)=0$.)
By replacing $(A,B)$ with an $\mathrm{SL}_3(K)$-translate, we may assume that
$\mathbb{P}^1_K$ is the subspace $(\ast, \ast, 0)$, and we may hence assume that
$a_{12}=a_{22}=b_{11}=b_{12}=0$.
Since $a_{11}$ and $b_{22}$ are nonzero, we may use another
$\mathrm{SL}_3(K)$-transformation to further assume that
$a_{13}=b_{23}=0$.
Then we have
\[
a/4 = \det(A) = -a_{11} a_{23}^2
\quad\text{and}\quad
-d/4 = \det(B) = -b_{22} b_{13}^2.
\]
Thus, with another $\mathrm{SL}_3(K)$-transformation, we can ensure that
$a_{11}=-a$, $b_{22}=d$, and $a_{23}=b_{13}=\tfrac12$.
Finally, the coefficients $a_{33}$ and $b_{33}$ are uniquely determined by the values of $b$ and $c$. The lemma follows.
\end{proof}

\end{document}